\documentclass[conference]{IEEEtran}
\IEEEoverridecommandlockouts
\usepackage[hidelinks]{hyperref}
\usepackage[most]{tcolorbox}
\usepackage{amssymb}
\usepackage{amsmath}
\usepackage{booktabs}
\usepackage{multirow}
\usepackage{graphicx} 
\usepackage{pgfplots}
\usepackage{tikz}
\pgfplotsset{compat=1.18}

\def\BibTeX{{\rm B\kern-.05em{\sc i\kern-.025em b}\kern-.08em
    T\kern-.1667em\lower.7ex\hbox{E}\kern-.125emX}}

\expandafter\def\expandafter\UrlBreaks\expandafter{\UrlBreaks\do\/\do\*\do\-\do\~\do\'\do\"\do\-}

\title{Neurosymbolic Retrievers for Retrieval-augmented Generation
\thanks{This work is supported by a UMBC Faculty Startup Award and a gift from NeuralNest LLC. The opinions expressed are those of the authors and do not necessarily reflect the views of UMBC or NeuralNest.}
}

\author{\IEEEauthorblockN{Yash Saxena}
\IEEEauthorblockA{\textit{Dept. of CSEE} \\
\textit{University of Maryland}\\
Baltimore County, Maryland, USA\\
ysaxena1@umbc.edu}
\and
\IEEEauthorblockN{Manas Gaur}
\IEEEauthorblockA{\textit{Dept. of CSEE} \\
\textit{University of Maryland}\\
Baltimore County, MD, USA\\
manas@umbc.edu}
}

\begin{document}
\maketitle
\begin{abstract}
Retrieval Augmented Generation (RAG) has made significant strides in overcoming key limitations of large language models, such as hallucination, lack of contextual grounding, and issues with transparency. However, traditional RAG systems consist of three interconnected neural components—the retriever, re-ranker, and generator—whose internal reasoning processes remain opaque. This lack of transparency complicates interpretability, hinders debugging efforts, and erodes trust, especially in high-stakes domains where clear decision-making is essential. To address these challenges, we introduce the concept of Neurosymbolic RAG, which integrates symbolic reasoning using a knowledge graph with neural retrieval techniques. This new framework aims to answer two primary questions: (a) Can retrievers provide a clear and interpretable basis for document selection? (b) Can symbolic knowledge enhance the clarity of the retrieval process? We propose three methods to improve this integration. First is \textbf{MAR} (Knowledge Modulation Aligned Retrieval)  that employs modulation networks to refine query embeddings using interpretable symbolic features, thereby making document matching more explicit. Second, \textbf{KG-Path RAG} enhances queries by traversing knowledge graphs to improve overall retrieval quality and interpretability. Lastly, Process Knowledge-infused RAG utilizes domain-specific tools to reorder retrieved content based on validated workflows. Preliminary results from mental health risk assessment tasks indicate that this neurosymbolic approach enhances both transparency and overall performance.
\end{abstract}


\section{Introduction}
\begin{quote}
    As machines learn, they may develop unforeseen strategies at rates that baffle their programmers--- \textit{Norbert Wiener (1960)} 
\end{quote}

This observation captures a central challenge in contemporary AI: the need for symbolic grounding, especially along the procedural grounding dimension, to ensure that machine outputs remain aligned with explicit, auditable reasoning processes \cite{10.1109/MIS.2024.3366669}. Although generative AI systems based on large language models (LLMs) now underpin applications in scientific discovery, healthcare, cybersecurity, and law, their fluency masks persistent shortcomings. LLMs routinely hallucinate facts , struggle to access current or specialized knowledge, lack transparent attribution, and operate without structured mechanisms for uncertainty detection or domain-specific procedural reasoning. These limitations constrain their reliability in high-stakes decision-making contexts.

Retrieval-Augmented Generation (RAG) has emerged as a promising remedy by conditioning generation on retrieved evidence. Its canonical retriever–re-ranker–generator pipeline enables temporal updates, factual grounding, personalization, and explicit citation, supporting use cases ranging from enterprise knowledge management to clinical support \cite{tilwani2024neurosymbolic}. Yet despite these strengths, RAG systems inherit a profound interpretability (ante-hoc and post-hoc) deficit: each component operates as a neural closed box. Developers cannot determine why particular documents were retrieved or ranked; users cannot verify which passages shaped the model’s answer; and domain specialists cannot confirm adherence to established workflows. Consequently, RAG’s nominal transparency by providing sources fails to guarantee trustworthy reasoning. 

In a clinical‐decision‐support context, a system might retrieve and present appropriate clinical guidelines, yet still fail by violating the correct sequence of care. For instance, first, a screening tool flags a potential concern; next, a clinical consultation obtains more context; then, a risk assessment identifies high-risk patients; followed by a diagnostic interview to apply formal criteria; and finally, the patient receives a guided intervention such as Cognitive Behavioural Therapy (CBT). If the system presents these steps out of order (e.g., recommending CBT before assessing risk or diagnosis), the result appears evidence-based yet mis-implements the workflow. Research supports such stepped workflows: systematic screening tools should trigger deeper assessment when needed. In turn, the formal diagnostic interview and treatment planning follow assessment and risk stratification. Thus, while accessing guideline content (\textit{retrieval}) is necessary, it is not sufficient: correct ordering, branching, and contextual logic remain critical to effective, safe care \cite{Cantrell2024RiskAssessment}.

Addressing this opacity requires a shift from post hoc explanation to inherently interpretable architectures. Neurosymbolic integration provides such a foundation by embedding structured symbolic representations directly into the retrieval pipeline. Knowledge graphs can reveal explicit semantic traversal paths, symbolic features can transparently modulate neural embeddings, and procedural instruments can enforce domain-appropriate reasoning steps. These elements transform retrieval from an opaque similarity search into an accountable, workflow-aligned reasoning process. By unifying neural flexibility with symbolic structure, neurosymbolic RAG offers a path toward verifiable, interpretable systems capable of meeting the demands of safety-critical domains.

This work introduces Neurosymbolic RAG, a framework that tightly couples symbolic reasoning with neural retrieval to address three fundamental research questions that challenge the closed box paradigm: 

\textbf{RQ1:} Can retrievers be made interpretable through symbolic feature accumulation? Traditional dense retrievers encode queries and documents into opaque embedding spaces, providing no insight into matching criteria. We \textit{theoretically} investigate whether retrieval can be augmented with symbolic features—accumulated across conversational context, that explicitly modulate embeddings to provide transparent justifications for document (or chunk) selection. \textit{(Section 2)}

\textbf{RQ2:} Can knowledge graphs simultaneously improve and explain retrieval? Structured knowledge offers both semantic enrichment and interpretable reasoning paths. We examine whether integrating knowledge graphs with neural retrieval via query enrichment and graph-based ranking can reveal the logical chains connecting queries to evidence while improving retrieval quality via semantically informed traversal paths. \textit{(Section 3)}

\textbf{RQ3:} Can procedural knowledge guide retrieval to follow domain-specific workflows? Beyond semantic relevance, many domains require adherence to validated assessment protocols. We explore whether incorporating structured procedural instruments—such as clinical questionnaires—can reorder retrieved evidence to follow the same stepwise reasoning processes used by domain experts. (Section 4)

\section{Related Work}

Recently, there have been significant efforts into neurosymbolic retrieval-augmented generation, such as ArgRAG and SymRAG \cite{zhu2025argrag, hakim2025symrag}. ArgRAG introduces a training-free neurosymbolic framework that combines RAG with Quantitative Bipolar Argumentation Framework (QBAF) for robust, explainable, and contestable fact verification and scientific feasibility, modeling support and attack relations between claims and retrieved evidence to compute final argument strengths. SymRAG presents a neuro-symbolic framework that introduces adaptive query routing based on real-time complexity and system load assessments, dynamically selecting symbolic, neural, or hybrid processing paths to align resource use with query demands while achieving high accuracy with significantly lower CPU utilization. 

However, these approaches primarily address post-hoc explainability through argumentation frameworks or computational efficiency through adaptive routing, but do not fundamentally transform the retrieval process itself to be inherently interpretable from the ground up. Furthermore, while ArgRAG provides faithful explanations for fact verification tasks, it treats each retrieved document chunk as a single argument without considering fine-grained argument extraction or the sequential ordering requirements imposed by domain-specific procedural workflows. Similarly, SymRAG optimizes resource allocation across query types but lacks mechanisms to enforce adherence to validated assessment protocols or established reasoning sequences that are critical in high-stakes domains such as clinical decision-support, where correct procedural ordering (e.g., screening before diagnosis, risk assessment before intervention) is essential for safe and effective care. 

RuleRAG introduces symbolic rules as demonstrations for in-context learning to guide retrievers toward logically related documents and generators toward rule-attributed answers, achieving 89.2\% improvement in Recall@10 . GraphRAG uses LLM-generated knowledge graphs to extract entity relationships and create community summaries, enabling hierarchical reasoning that substantially outperforms baseline RAG on global queries requiring dataset-wide understanding. 

While RuleRAG \cite{chen2024rulerag}, and GraphRAG \cite{edge2024local}, enhance RAG through symbolic reasoning and knowledge graphs, they differ fundamentally from MAR, KG-Path RAG, and Proknow-RAG in both architectural integration and interpretability mechanisms. RuleRAG and GraphRAG augment retrieval with external symbolic structures but maintain decoupling between symbolic reasoning and neural retrieval. GraphRAG uses KGs primarily for query enrichment and post-retrieval organization, while the actual retrieval remains in vector space. KG-Path RAG addresses this limitation through joint optimization, treating KG traversal and retrieval as a unified process where graph-based ranking losses (analogous to PageRank) iteratively refine retrieval distributions, achieving query enrichment more efficiently while providing explicit path-based provenance. MAR differs further by making embeddings themselves interpretable through modulation networks that project symbolic features directly into the neural representation. Finally, while these approaches improve semantic grounding and logical consistency, Proknow-RAG uniquely enforces procedural correctness by leveraging domain-specific instruments to reorder evidence according to validated clinical workflows, ensuring not just relevance but adherence to established assessment sequences critical in high-stakes domains.

\section{Knowledge Modulation-aligned Retrieval}
Traditional Dense Passage Retrieval encodes queries and documents into opaque vector spaces where we don't understand why certain documents match certain queries—we only see that their cosine similarity is high. Instead of directly using raw neural embeddings, we introduce modulation networks that adjust embeddings based on interpretable, domain-specific factors. The query path processes text through a base encoder followed by a modulation network to produce a modulated embedding, while the document path follows an identical structure. The modulation networks decompose the embedding space into interpretable dimensions that can be visualized, explained, and controlled. Think of it as adding "semantic dials" that explicitly control which aspects of meaning are emphasized.

\noindent \textit{\textbf{Clinical Conversation Scenario:}} Consider the following natural clinical conversation, where the task for Neurosymbolic RAG is to identify possible mental health and psychological risk factor assessments that can be populated from an active conversation between patient and clinician. The clinician begins by asking, "Thanks for coming in today. How have you been feeling lately?" The patient responds: "Honestly, not great. I've been really down, you know? Nothing seems fun anymore, not even things I used to love." The patient continues: "And I can't sleep. I lie awake for hours every night, just... thinking. Remembering things." When the clinician probes further about these memories, the patient discloses: "Stuff from when I was a kid. My dad... he had a bad temper. There was a lot of yelling, sometimes worse. I don't know why it's all coming back now." The patient concludes: "I'm exhausted all the time, but my brain won't shut off at night. Then I'm useless during the day."

The challenge of Multi-Turn Clinical Conversations: As the conversation unfolds, the purpose of a Neurosymbolic system is to continuously process each utterance, extract symbolic features that build a progressive, rich clinical picture for proper assessment. Traditional closed box retrieval systems struggle with natural conversations because they lack the structured knowledge to integrate information across multiple turns and understand causal relationships between disclosed facts. A standard neural system might retrieve depression articles when the patient mentions low mood, then separately retrieve trauma articles when abuse is mentioned, and sleep articles when insomnia comes up, but it wouldn't understand that these form a coherent clinical pattern requiring integrated assessment and treatment. The neurosymbolic modulation approach succeeds because the symbolic knowledge structures explicitly encode these relationships: 

\begin{tcolorbox}[
    colback=blue!5,
    colframe=purple!60!black,
    fonttitle=\bfseries,
    boxrule=0.8pt,
    arc=2mm,
    left=1.5mm,
    right=1.5mm,
    top=1mm,
    bottom=1mm
]
ACE exposure $\rightarrow$ increases risk for $\rightarrow$ Sleep Disorder $\rightarrow$ maintains $\rightarrow$ Depression $\rightarrow$ requires $\rightarrow$ Trauma-informed Treatment. \\ 
\textit{ACE: Adverse Childhood Experiences.}
\end{tcolorbox}

The neural component provides flexible pattern matching across diverse ways patients describe experiences ("brain won't shut off" = hyperarousal; "nothing seems fun" = anhedonia), while the symbolic component ensures these patterns are interpreted within established clinical questionnaires.

Traditional closed box Retrieval at turn \textit{t} operates independently for each utterance. The scoring function can be expressed as: 
\[\mathrm{score}(\mathrm{query}^t, \mathrm{doc}) = \mathrm{cosine}(e_q^{t}, e_d)\]

Each utterance is processed independently. If the patient mentions depression at turn 1, trauma at turn 2, and sleep problems at turn 3, the system performs three separate retrievals with no connection between them.

\noindent \textit{\textbf{Neurosymbolic Retrieval with Accumulation:}} Neurosymbolic retrieval addresses this limitation through feature accumulation across conversation turns. The accumulation process builds a progressively richer representation of the clinical picture as the conversation unfolds. At each turn t, the system maintains an accumulated set of symbolic features $\varphi(t)$ that combines all previously extracted features with newly identified ones. This accumulation is formally expressed as:

\begin{align*}
\varphi(t) &= \varphi(t-1) \cup \varphi_{\text{new}}(t) \\
e'_{q}{}^{t} &= e_{q}^{t} + \alpha^{t} \cdot W_{q} \cdot \varphi(t)
\end{align*}

Here, $\varphi(t)$ accumulates symbolic features across turns, building a comprehensive clinical picture. At turn 1 ($t = 1$), when the patient states ``I've been feeling really down lately,'' the system extracts the symbolic feature \{depressed\_mood\}, so $\varphi(1) = \{\text{depressed mood}\}$. At turn 2 ($t = 2$), the patient mentions ``Nothing seems fun anymore,'' leading to the extraction of new features $\varphi_{\text{new}}(2) = \{\text{anhedonia}\}$. The accumulated feature set now becomes $\varphi(2) = \varphi(1) \cup \varphi_{\text{new}}(2) = \{\text{depressed mood}, \text{anhedonia}\}$. By turn 3 ($t = 3$), when the patient discloses ``I can't sleep. I lie awake thinking about my dad's temper when I was a kid,'' the system extracts $\varphi_{\text{new}}(3) = \{\text{chronic\_insomnia}, \text{rumination}, \text{ACE disclosure}, \\
\text{childhood abuse}\}$. The complete accumulated feature set becomes $\varphi(3) = \{\text{depressed mood}, \text{anhedonia}, \\ \text{chronic insomnia}, \text{rumination}, \text{ACE disclosure}, \\ \text{childhood abuse}\}$.

The modulation strength $\alpha^t$ is computed dynamically based on clinical complexity. As the feature set grows and reveals more interconnected symptoms, the system increases reliance on structured clinical knowledge. The modulation strength is determined through a sigmoid function:
\[
\alpha^t = \sigma(k \cdot \text{complexity}(\varphi(t)))
\]

Where $\sigma$ is the sigmoid function, and $k$ controls sensitivity. The complexity metric itself is computed from three components that capture different aspects of the clinical presentation:
\begin{equation*}
\text{complexity}(\varphi(t)) = |\varphi(t)| + \sum_{i,j \in \varphi(t)} w_{ij} + \sum_{i \in \varphi(t)} r_i
\end{equation*}

The first term $|\varphi(t)|$ represents the symptom count---the total number of distinct clinical features identified in a patient-clinician conversation. The second term captures graph connectivity, summing the connection weights $w_{ij}$ between features $i$ and $j$ within the clinical knowledge graph. The third term incorporates risk weighting, summing the clinical severity scores $r_i$ for each feature. Together, these components provide a nuanced measure of how complex and interconnected the clinical presentation has become.

The modulated query embedding $e'_q(t)$ is generated by shifting the base neural embedding $e_q(t)$ through the modulation term $W_q \cdot \varphi(t)$, which projects the accumulated symbolic features into the embedding space. The projection matrix $W_q$ has been learned during training to map each symbolic clinical concept to an appropriate direction in the high-dimensional embedding space. Documents in the database are similarly modulated based on their extracted symbolic features $\varphi_d$. The document modulation follows the same principle:
\begin{equation*}
e'_d = e_d + \beta \cdot W_d \cdot \varphi_d
\end{equation*}

For example, consider a document titled ``Integrated Assessment for Trauma-Related Depression'' that has been annotated with symbolic features $\varphi_d$ = \{trauma protocol, sleep assessment, Major Depressive Disorder criteria, ACE screening\}. This document's embedding is pre-computed and modulated using these features, positioning it in the embedding space where queries with similar feature combinations will find it through high cosine similarity. The retrieval score between a query at turn $t$ and a candidate document is computed as the cosine similarity between their modulated embeddings:
\begin{equation*}
\text{score}(\text{query}^t, \text{doc}) = \text{cosine}(e'_{q}{}^{t}, e'_d)
\end{equation*}

Finally, the system ranks all documents in the database according to their scores and retrieves the top-$k$ most relevant documents. This ranking can be expressed as retrieving the set of documents with the highest scores: $\text{Retrieved} = \text{top-}k\{\text{doc}_i: \text{score}(\text{query}^t, \text{doc}_i)\}$. Through this neurosymbolic approach, the system successfully identifies that the patient's disclosure across three turns represents an integrated clinical syndrome requiring comprehensive trauma-informed assessment, rather than three separate issues to be addressed independently.

\section{KG-Path RAG}

Standard vector-based RAG architectures typically do not propagate explicit retrieval-path information (e.g., document lineage or multi-hop transitions), which limits the interpretability and fine-grained traceability of retrieval decisions. Methods such as GraphRAG mitigate this limitation by inducing a knowledge graph over the corpus—where nodes represent entities or text units and edges encode semantic relationships—and then performing path-aware retrieval over this structure, yielding more coherent and interpretable evidence chains for complex queries \cite{edge2024local}. However, much of the recent work on provenance and traceability in RAG, such as citation-highlighting and interactive visual evaluation tools like RAGTrace, has primarily focused on analyzing and visualizing retrieval behavior \cite{cheng2025ragtrace}, rather than on architectures that incorporate knowledge-graph structures to iteratively enrich user queries and enhance ranking quality via graph-aware retrieval. In parallel, knowledge-graph-enhanced RAG and query-refinement methods have emerged as separate lines of work, aiming to leverage structured graphs or iterative query rewriting to improve retrieval quality and ranking, yet there remains relatively little research that tightly integrates these directions into a single, knowledge-graph–infused, path-aware RAG framework that both enriches queries and optimizes document ranking using explicit provenance signals. One line of work that attempts to address the lack of query-aware graph optimization in GraphRAG is G-Retriever \cite{he2024gretriever}, which conditions retrieval on subgraphs extracted through prize-collecting Steiner tree (PCST) optimization. This approach can yield more precise query-specific subgraphs than heuristic GraphRAG traversals. However, G-Retriever requires jointly fine-tuning the model on both the GraphRAG-induced graph structure and the query distribution, introducing substantial additional complexity. Moreover, because the model learns implicit semantic associations during fine-tuning, explicit control over graph-level semantics, and thus, the interpretability of the selected subgraph remains limited.

\begin{figure*}
    \centering
    \includegraphics[width=\textwidth]{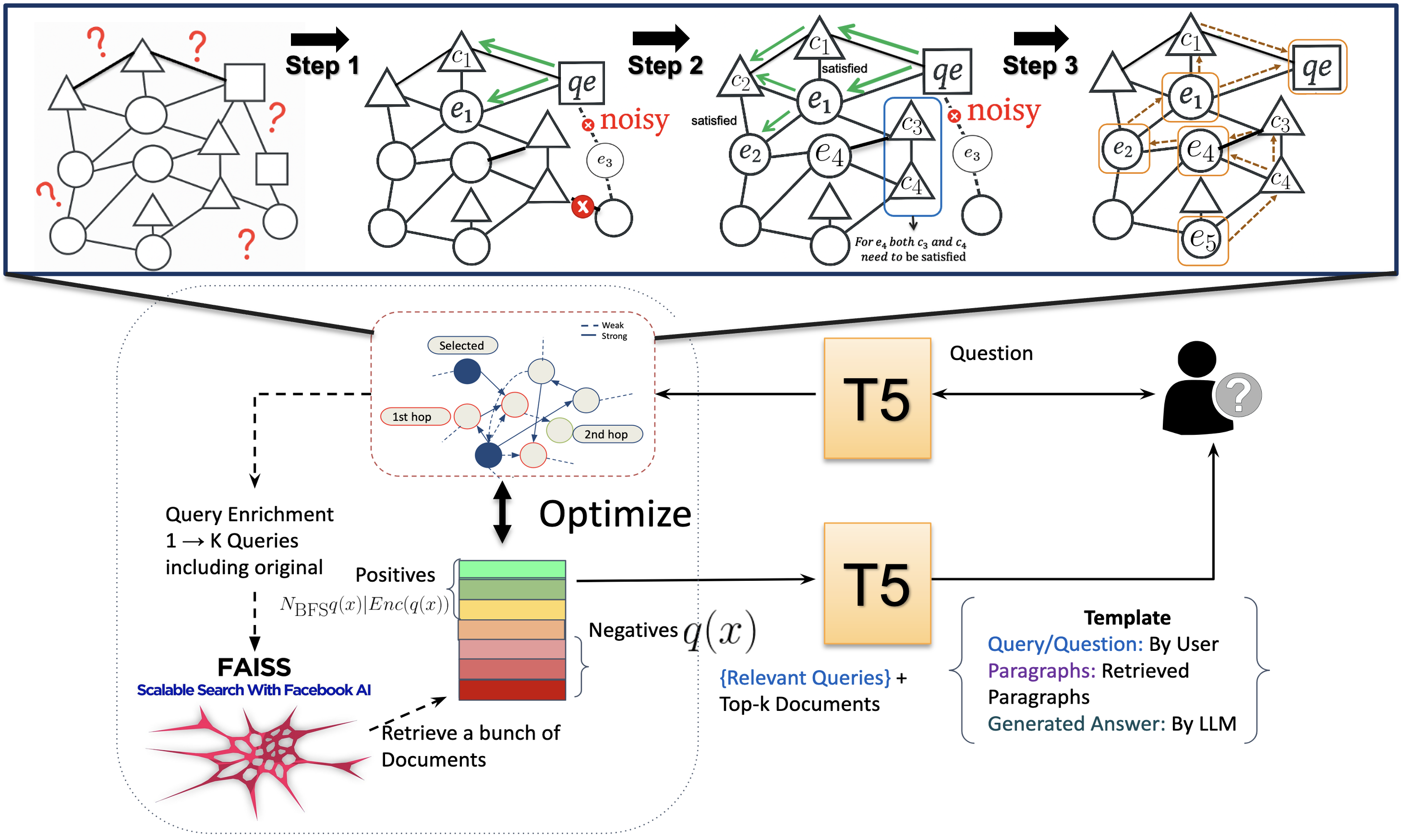}
    \caption{Overview of KG-Path RAG training and inference pipeline. \textbf{Top}: Graph-based ranking loss training process across three steps. Initially, the knowledge graph contains uncertain relationships (Step 1). During training, the system learns to identify satisfied edges (green) and filter noisy connections (red X), progressively building reliable reasoning paths through entities ($e_1, e_2, e_3, e_4$) to satisfy query constraints ($qe$) (Steps 2-3). \textbf{Bottom}: Inference pipeline showing query enrichment from a single user question to $K$ queries (including the original), followed by multi-hop graph traversal with BFS (Breadth First Search)-guided exploration that distinguishes strong and weak connections across 1st and 2nd hop neighbors. FAISS (Facebook AI Semantic Search) retrieves candidate documents while the ranking function $q(x)$ separates positives from negatives. The optimized retrieval results combine relevant queries and top-$k$ documents, which are structured through a template (Query/Question by User, Retrieved Paragraphs, Generated Answer by LLM) and fed to the T5 model for final answer generation. This joint optimization of KG traversal and retrieval enables explicit path-based provenance while maintaining semantic grounding. The bi-directional arrow between user and T5 represents a RAG-disabled case, similar to MentalLLAMA, a fine-tuned model answering mental health queries.}
    \label{fig:1}
\end{figure*}

To address the limitations of standard RAG and the gaps identified in graph-augmented retrieval, we introduce two complementary improvements: (1) the integration of a knowledge graph (KG) to support query enrichment and semantic-preserving retrieval, and (2) explicit mechanisms for interpretability and traceability throughout the retrieval pipeline. Incorporating a KG adds three key capabilities that do not arise naturally in vanilla RAG: query enrichment, logical coherence, and semantic relatedness in retrieval. In our framework, the user query is first mapped onto the KG, and one-hop or two-hop traversals are used to surface conceptually related nodes. The original query, together with these KG-derived concepts, is then provided to a language model to generate an enriched query. This enriched query is used to contextualize downstream retrieval via standard vector-based RAG, thereby improving the quality and relevance of the retrieved evidence. The resulting generation quality can be systematically evaluated using a language model fine-tuned on natural-language inference to measure logical coherence, and another model fine-tuned on semantic similarity tasks to assess semantic relatedness.

A remaining limitation of this approach is that KG traversal is primarily used for query enrichment, while the actual retrieval step remains decoupled from the KG structure. As a result, although the enriched query preserves a clear provenance trail to the KG, the final retrieval can still occur out of context, since the vector-space retriever is not directly conditioned on the KG traversal or its evolving semantic structure. This motivates treating KG traversal, query enrichment, and retrieval as a joint optimization problem, rather than sequential, disjoint components. Under such a formulation, the retrieval distribution would be progressively refined as KG traversal unfolds, allowing graph-derived evidence to dynamically shape ranking decisions. This perspective naturally aligns with graph-based ranking losses, analogous to PageRank-style propagation, in which node importance is iteratively updated based on both structural connectivity and query-specific signals, ultimately enabling a fully graph-aware, semantically grounded retrieval process. We refer to this as KG-Path RAG and it can be formulated as follows: 

\paragraph{1. Query-to-Graph Mapping \& Traversal:}
Given a user query $q_e$, map it to initial nodes in knowledge graph $G = (V, E)$:
\[
\mathcal{N}(q_e) = \{e_i \in E : \text{sim}(q_e, e_i) > \tau\}
\]
Perform $k$-hop traversal to gather related concepts:
\[
\mathcal{N}_k(q_e) = \bigcup_{i=1}^{k} \text{BFS}_i(\mathcal{N}(q_e), KG)
\]

\paragraph{2. Query Enrichment:}
Generate enriched query using LLM (e.g., masked language model or autoregressive model) conditioned on original query and KG-derived concepts:
\[
q_{\text{enriched}} = \text{LLM}(q_e \cup \mathcal{N}_k(q_e))
\]

\paragraph{3. Graph-Based Ranking Function:}
For each document $d_j$, compute retrieval score incorporating both semantic similarity and graph connectivity:
\begin{align*}
\text{score}(q_{\text{enriched}}, d_j) &= \alpha \cdot \text{cosine}(\text{Enc}(q_{\text{enriched}}), \text{Enc}(d_j)) \\
&\quad + (1-\alpha) \cdot \text{PR}(d_j, KG)
\end{align*}
where PR (PageRank) iteratively refines node importance:
\[
\text{PR}(e_i) = \frac{1-\gamma}{|E|} + \gamma \sum_{e_j \in \text{In}(e_i)} \frac{\text{PR}(e_j)}{|\text{Out}(e_j)|}
\]

\paragraph{4. Joint Optimization Objective:}
Minimize ranking loss that combines retrieval accuracy with graph structure:
\begin{align}
\mathcal{L} &= \mathcal{L}_{\text{retrieval}} + \beta \cdot \mathcal{L}_{\text{graph-rank}} \\
\mathcal{L}_{\text{graph-rank}} &= -\sum_{(q,d^+,d^-)} \log \frac{e^{\text{score}(q,d^+)}}{e^{\text{score}(q,d^+)} + e^{\text{score}(q,d^-)}}
\end{align}
where $d^+$ are relevant documents and $d^-$ are negatives based on graph-guided sampling. The hyperparameters $\gamma$ is set to 0.15 and $\beta$ is set to 0.40 to achieve results shown in Table \ref{tab:mental_health_performance}. 

\paragraph{5. Final Retrieval:}
\[
\text{Retrieved} = \text{top-}k\{d_j : \text{score}(q_{\text{enriched}}, d_j)\}
\]

Figure \ref{fig:1} describes the process of training interpretable and traceable retrievers using a graph-based ranking loss and illustrates how the KG is traversed during retriever training, along with the resultant ``path'' that can be used for reasoning behind the retrieval and the downstream-generated answers. 

\section{Process Knowledge-infused RAG (Proknow-RAG)}

While logical coherence and semantic relatedness, enforced via a knowledge graph and appropriate scoring models, help ensure that retrieved documents are locally consistent and topically aligned, they do not by themselves guarantee that the resulting document sequence adheres to an external, domain-specific process. In many domains, such as mental health, there exist well-established procedural instruments, e.g., the Patient Health Questionnaire (PHQ-9) for depression \cite{kroenke2001phq9} and the Generalized Anxiety Disorder scale (GAD-7) for anxiety \cite{spitzer2006gad7}—that encode an implicit reasoning and assessment workflow followed by clinicians. In KG-Path RAG, these questionnaires can be treated as structured procedural knowledge, providing an external template to order retrieved passages so that the information is presented in the same stepwise fashion as a clinician's assessment. This is analogous, at a domain level, to how ReAct-style prompting \cite{yao2022react} and how process reward models in the LLM literature \cite{lightman2023letverify} supervise and reward the quality of the reasoning process rather than only the final answer; here, the questionnaire defines a human-designed reasoning process, and retrieval is guided to follow that process in a clinically meaningful order.

Methodologically, we characterize this as process-knowledge–infused RAG, since the approach operates at the intersection of KG-guided retrieval and the subsequent procedural reordering of passages based on domain-specific process knowledge encoded in the clinical questionnaire \cite{sheth2022process}. In this formulation, the knowledge graph provides semantic and logical coherence, while the questionnaire acts as an external procedural scaffold that constrains the ordering of evidence, enabling the retrieval pipeline to follow the same stepwise assessment process used by domain experts.

Imagine a scenario, the AI assistant is conducting a mental-health follow-up session. As the person speaks, the assistant identifies important clinical terms, such as symptoms or emotional states, using a mental-health knowledge graph, which organizes how symptoms, conditions, and questionnaires relate to each other \cite{dalal2024cross}. The system then retrieves the person's past information and asks focused follow-up questions like, ``You previously reported X. Are you still experiencing it?'' This step depends on accurate retrieval of the user's history and relevant clinical concepts, which is handled by the KG-Path RAG's retrieval framework.

\begin{table*}[!htbp]
\centering
\caption{Performance comparison of KG-Path RAG, Proknow-RAG, and MentalLLAMA-33B (Baseline) across multiple mental health detection tasks on the IMHI dataset. Results are reported in terms of accuracy (Acc), F1 score, precision (Prec), and recall (Rec). Bold values indicate the best performance for each metric within a given task. The T5 model used in KG-Path RAG and Proknow-RAG has 3 billion parameters. }
\label{tab:mental_health_performance}
\resizebox{\textwidth}{!}{%
\begin{tabular}{@{}lc|cccc|cccc|cccc@{}}
\toprule
\multirow{2}{*}{\textbf{Mental Health Task}} & \multirow{2}{*}{\textbf{Samples}} & \multicolumn{4}{c|}{\textbf{KG-Path RAG}} & \multicolumn{4}{c|}{\textbf{Proknow-RAG}} & \multicolumn{4}{c}{\textbf{MentalLLAMA-33B}} \\
\cmidrule(lr){3-6} \cmidrule(lr){7-10} \cmidrule(lr){11-14}
& & \textbf{Acc} & \textbf{F1} & \textbf{Prec} & \textbf{Rec} & \textbf{Acc} & \textbf{F1} & \textbf{Prec} & \textbf{Rec} & \textbf{Acc} & \textbf{F1} & \textbf{Prec} & \textbf{Rec} \\
\midrule
Depression Detection & 2,150 & \textbf{84.7\%} & \textbf{0.832} & \textbf{0.851} & \textbf{0.814} & 82.3\% & 0.807 & 0.831 & 0.784 & 83.4\% & 0.821 & 0.838 & 0.805 \\
Anxiety Detection & 1,890 & 78.9\% & 0.773 & 0.796 & 0.751 & 75.6\% & 0.738 & 0.771 & 0.707 & \textbf{79.8\%} & \textbf{0.784} & \textbf{0.802} & \textbf{0.767} \\
PTSD Detection & 1,456 & \textbf{82.1\%} & \textbf{0.809} & \textbf{0.827} & \textbf{0.792} & 79.4\% & 0.777 & 0.801 & 0.755 & 81.2\% & 0.798 & 0.818 & 0.779 \\
Suicide Risk & 987 & 76.3\% & 0.745 & 0.781 & 0.712 & \textbf{89.1\%} & \textbf{0.885} & \textbf{0.897} & \textbf{0.873} & 75.6\% & 0.742 & 0.769 & 0.716 \\
Stress Analysis & 2,341 & \textbf{81.2\%} & \textbf{0.798} & \textbf{0.815} & \textbf{0.782} & 77.8\% & 0.761 & 0.785 & 0.738 & 80.1\% & 0.789 & 0.806 & 0.773 \\
Bipolar Detection & 678 & 73.4\% & 0.719 & 0.748 & 0.692 & 70.1\% & 0.683 & 0.719 & 0.649 & \textbf{74.3\%} & \textbf{0.728} & \textbf{0.751} & \textbf{0.706} \\
Eating Disorders & 543 & \textbf{69.8\%} & \textbf{0.681} & \textbf{0.715} & \textbf{0.649} & 66.5\% & 0.647 & 0.683 & 0.613 & 68.7\% & 0.671 & 0.698 & 0.646 \\
Self-harm Risk & 834 & 77.1\% & 0.756 & 0.778 & 0.726 & \textbf{83.2\%} & \textbf{0.819} & \textbf{0.845} & \textbf{0.794} & 77.8\% & 0.764 & 0.785 & 0.744 \\
\bottomrule
\end{tabular}%
}
\end{table*}

When a symptom is confirmed or a new one is reported, the assistant uses the knowledge graph to determine which standardized questionnaires apply (e.g., PHQ-9 for depression or GAD-7 for anxiety). Selecting the right questionnaire is a retrieval problem, because the system must match user-reported symptoms to the appropriate assessment tool. The knowledge graph ensures this matching is clinically valid.

After selecting the questionnaire, the system uses its structured items to determine the order of the questions it asks. This keeps the conversation aligned with the clinical workflow followed by human practitioners. Organizing information and questions in this structured way is supported by KG-Path RAG, which helps the system follow meaningful paths in the knowledge graph and maintain a coherent sequence during retrieval and questioning.

Across the full benchmark, KG-Path RAG and Proknow-RAG consistently deliver stronger, more clinically aligned performance than MentalLLAMA-33B (refer table \ref{tab:mental_health_performance}). KG-Path RAG excels in tasks that depend on precise symptom interpretation. It achieves the best results in Depression Detection (84.7\%) and shows strong performance in Stress Analysis (81.2\%). These gains stem from its knowledge-graph–guided retrieval, which helps the system understand how symptoms cluster, relate, and differentiate across conditions. This deeper semantic grounding allows KG-Path RAG to retrieve more clinically relevant evidence than a pure language-model approach. Proknow-RAG shines when the task requires structured clinical reasoning. It delivers the highest score in Suicide Risk Detection (89.1\%) and remains strong in Self-harm Risk Detection (83.2\%). The questionnaire-based process layer enforces a validated clinical workflow, guiding the model to ask the right questions in the right order. This produces safer and more reliable assessments, especially in high-risk categories where procedural structure matters. MentalLLAMA-33B performs well in Anxiety Detection (79.8\%) and offers balanced results overall, with strong explanation-generation ability. However, it lacks the semantic scaffolding of KG-Path RAG and the procedural reasoning of Proknow-RAG, limiting its diagnostic precision in complex or high-risk assessments.

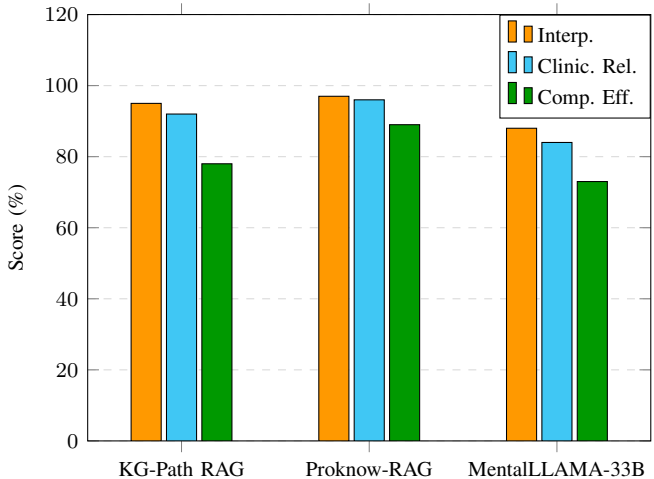
\begin{figure}[!htbp]
\centering
\begin{tikzpicture}
\begin{axis}[
    ybar,
    bar width=0.4cm,
    width=0.5\textwidth,
    height=0.4\textwidth,
    ylabel={Score (\%)},
    ylabel style={font=\footnotesize},
    ymin=0,
    ymax=120,
    xtick=data,
    symbolic x coords={KG-Path RAG, Proknow-RAG, MentalLLAMA-33B},
    xticklabel style={font=\footnotesize},
    yticklabel style={font=\footnotesize},
    legend style={
        at={(1.,1.)},
        anchor=north east,
        font=\footnotesize,
        draw=black,
        fill=white,
        cells={anchor=east}
    },
    ymajorgrids=true,
    grid style={dashed, gray!30},
    enlarge x limits=0.25,
    legend cell align={left},
]

\addplot[
    fill=orange!80!yellow,
    draw=black,
] coordinates {
    (KG-Path RAG, 95)
    (Proknow-RAG, 97)
    (MentalLLAMA-33B, 88)
};

\addplot[
    fill=cyan!60,
    draw=black,
] coordinates {
    (KG-Path RAG, 92)
    (Proknow-RAG, 96)
    (MentalLLAMA-33B, 84)
};

\addplot[
    fill=green!60!black,
    draw=black,
] coordinates {
    (KG-Path RAG, 78)
    (Proknow-RAG, 89)
    (MentalLLAMA-33B, 73)
};

\legend{Interp., Clinic. Rel., Comp. Eff.}

\end{axis}
\end{tikzpicture}
\caption{Human evaluation comparing KG-Path RAG, Proknow-RAG, and MentalLLAMA-33B on mental health detection tasks sourced from the IMHI dataset\cite{yang2024mentallama}. Three attributes were measured: Interpretability (Interp.), Clinical Relevance (Clinic. Rel.), and Computational Efficiency (Comp. Eff.). \textit{Interpretability} assesses whether retrieved chunks provide transparent reasoning for answer generation. \textit{Clinical Relevance} evaluates the presence and accuracy of clinical entities in retrieved chunks that align with the input query. \textit{Computational Efficiency} measures the time required to generate complete responses.} 
\label{fig:qualitative_metrics}
\end{figure}

\textbf{Expert Evaluation} with three Licensed Clinical Psychologists with Agreement of 75\%: Figure \ref{fig:qualitative_metrics} compares interpretability, clinical relevance, and computational efficiency across the three systems. Proknow-RAG performs best, scoring highest in interpretability, because the questions it generates are both clinically valid and informative for assessment. It also leads to clinical relevance, reflecting its ability to follow established diagnostic workflows. KG-Path RAG performs similarly well, supported by its knowledge graph grounding, which helps maintain clear, meaningful reasoning. MentalLLAMA-33B performs adequately but shows lower interpretability and clinical alignment. Proknow-RAG is also the most computationally efficient, making it the most practical and reliable option for real clinical decision-support scenarios.

\section{Conclusion and Future Directions}

This work demonstrates that neurosymbolic approaches can fundamentally reshape how clinical retrieval systems reason over multi-turn patient narratives. By embedding structured clinical knowledge directly into the retrieval process, these architectures move beyond surface-level pattern matching and instead support decisions that can be justified, interrogated, and aligned with clinical logic. The result is a retrieval paradigm in which system behavior is no longer a byproduct of opaque embedding geometries, but the consequence of explicit clinical constructs that remain visible throughout the reasoning pipeline. Across the designs explored, a consistent theme emerges: neurosymbolic systems provide a means to preserve the linguistic adaptability of neural models while grounding their behavior in the normative structure of clinical assessment. This hybridization is particularly important for conversations involving trauma, multi-symptom presentations, or evolving disclosure patterns. In these settings, retrieval quality is not defined solely by relevance, but by whether the system recognizes the clinical relationships implicit in the patient’s unfolding story. The architectures presented here show how symbolic structure can anchor that process, ensuring that patient narratives are interpreted cohesively rather than as isolated fragments.

Looking forward, several research directions are essential for advancing neurosymbolic retrieval toward real clinical deployment. First, systematic evaluation of transparency, evidentiary traceability, and alignment with clinical reasoning is required—especially in comparison to emerging agentic and closed box retrieval methods that lack explicit auditability. Second, expanding these systems to operate over multimodal sources, such as neuroimaging, physiological monitoring, or longitudinal patient records, raises the challenge of maintaining interpretability as data complexity grows. Third, establishing clinician-centered evaluation frameworks will be critical. Retrieval outputs must be assessed not merely for correctness, but also for their appropriateness within professional practice, consistency with diagnostic guidance, and safety within high-stakes decision contexts. Ultimately, the central question is whether neurosymbolic systems can meet the rigor demanded of tools intended for clinical workflows: grounding (relates to reliability), transparency, and the capacity to justify their recommendations in terms that clinicians can trust. The architectures introduced here represent an important step toward that goal, illustrating how symbolic structure and neural flexibility can be combined to support retrieval that is not only accurate but meaningfully aligned with clinical reasoning.

\bibliography{reference}
\bibliographystyle{ieeetr}

\end{document}